\pdfoutput=1

\documentclass[11pt]{article}

\usepackage[]{ACL2023}

\usepackage{times}
\usepackage{latexsym}
\usepackage[T1]{fontenc}

\usepackage[utf8]{inputenc}

\usepackage{microtype}

\usepackage{inconsolata}

\usepackage{amsmath} 
\usepackage{graphicx}
\usepackage{booktabs}
\graphicspath{ {figures/} }
\usepackage{siunitx}
\usepackage{enumitem}
\usepackage{subfig}
\usepackage{tabularx}
\usepackage{verbatim}
\usepackage{float}
\usepackage{kotex}

\title{Should Cross-Lingual AMR Parsing go Meta? An Empirical Assessment \\ of Meta-Learning and Joint Learning AMR Parsing}

\author{
 \textbf{Jeongwoo Kang\textsuperscript{1,2}} \quad
 \textbf{Maximin Coavoux\textsuperscript{1}} \quad
 \textbf{Cédric Lopez\textsuperscript{2}} \quad
 \textbf{Didier Schwab\textsuperscript{1}} \\
 \textsuperscript{1}Univ. Grenoble Alpes, CNRS, Grenoble INP,  LIG, 38000 Grenoble, France \\
 \textsuperscript{2}Emvista, Immeuble Le 610, 10 Rue Louis Breguet Bâtiment D, 34830 Jacou, France \\ 
\textsuperscript{1}\nolinkurl{{firstname}.{lastname}@univ-grenoble-alpes.fr} \\
\textsuperscript{2}\nolinkurl{{firstname}.{lastname}@emvista.com}
 }

\begin{document}
\maketitle
\begin{abstract}

Cross-lingual AMR parsing is the task of predicting AMR graphs in a target language when training data is available only in a source language. Due to the small size of AMR training data and evaluation data, cross-lingual AMR parsing has only been explored in a small set of languages such as English, Spanish, German, Chinese, and Italian. Taking inspiration from \citet{langedijk-etal-2022-meta}, who apply meta-learning to tackle cross-lingual syntactic parsing, we investigate the use of meta-learning for cross-lingual AMR parsing. We evaluate our models in $k$-shot scenarios (including 0-shot) and assess their effectiveness in Croatian, Farsi, Korean, Chinese, and French. Notably, Korean and Croatian test sets are developed as part of our work, based on the existing \textit{The Little Prince} English AMR corpus, and made publicly available. We empirically study our method by comparing it to classical joint learning. Our findings suggest that while the meta-learning model performs slightly better in 0-shot evaluation for certain languages, the performance gain is minimal or absent when $k$ is higher than 0. 

\end{abstract}

\section{Introduction}
Abstract Meaning Representation \cite[AMR]{Banarescu2013} represents the meaning of texts as rooted and directed acyclic graphs. AMR graphs capture the underlying semantics of input texts while abstracting away from their syntactic realizations. Nodes in AMR graphs are not explicitly mapped to their input token. Hence, it is an unanchored formalism. AMRs are widely used to enhance the capabilities of NLP systems such as question answering \cite{deng2022interpretable,kapanipathi-etal-2021-leveraging}, text summarization \cite{liao-etal-2018-abstract,liu-etal-2015-toward}, or human-robot interaction \cite{bonial-etal-2019-abstract,bonial-etal-2023-abstract}. 

\begin{figure}%
    \centering
    \subfloat[\centering AMR graph]{{\includegraphics[width=0.4\columnwidth]{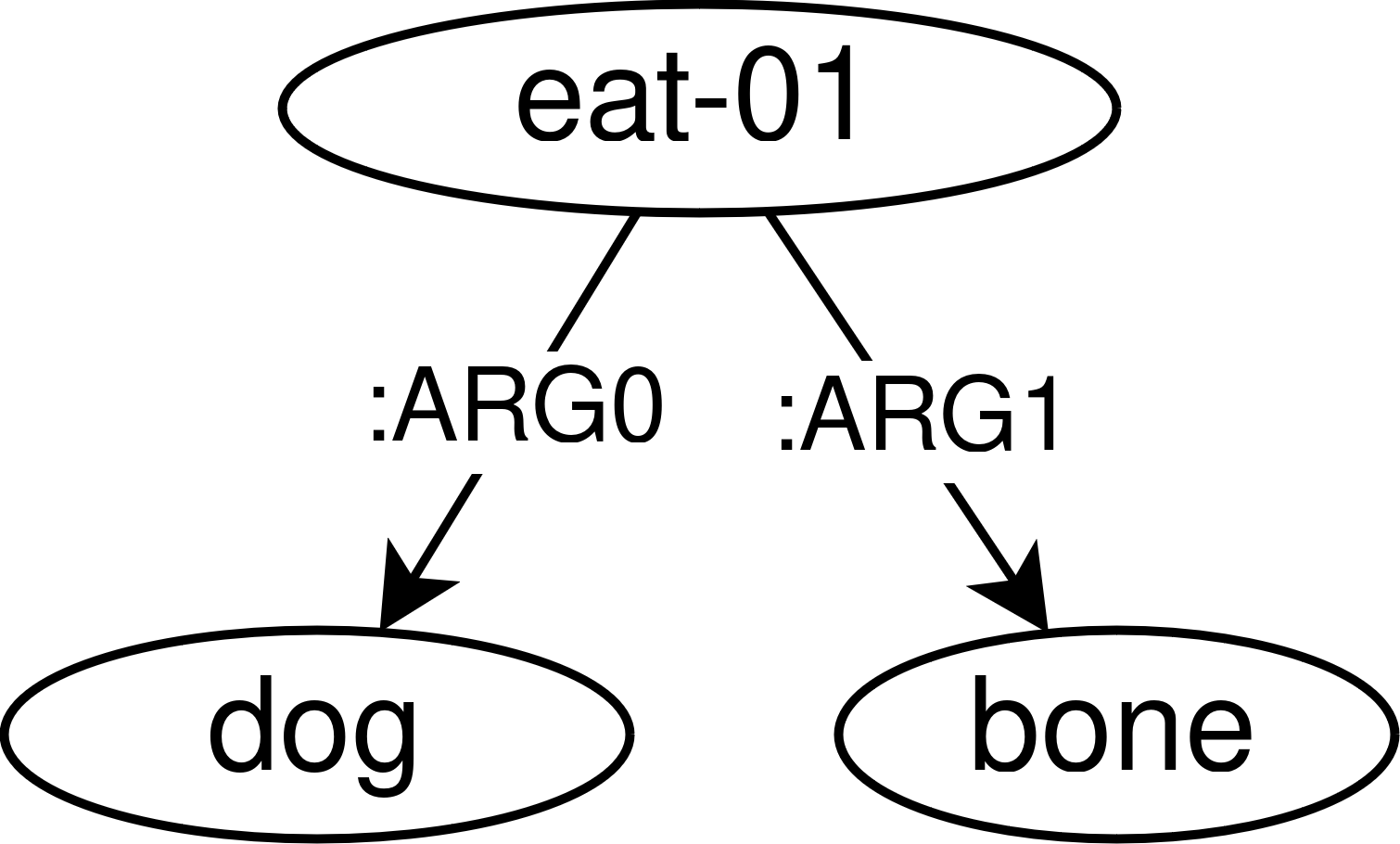} }}%
    \qquad
    \subfloat[\centering linearized AMR]
    {{\includegraphics[width=0.38\columnwidth]{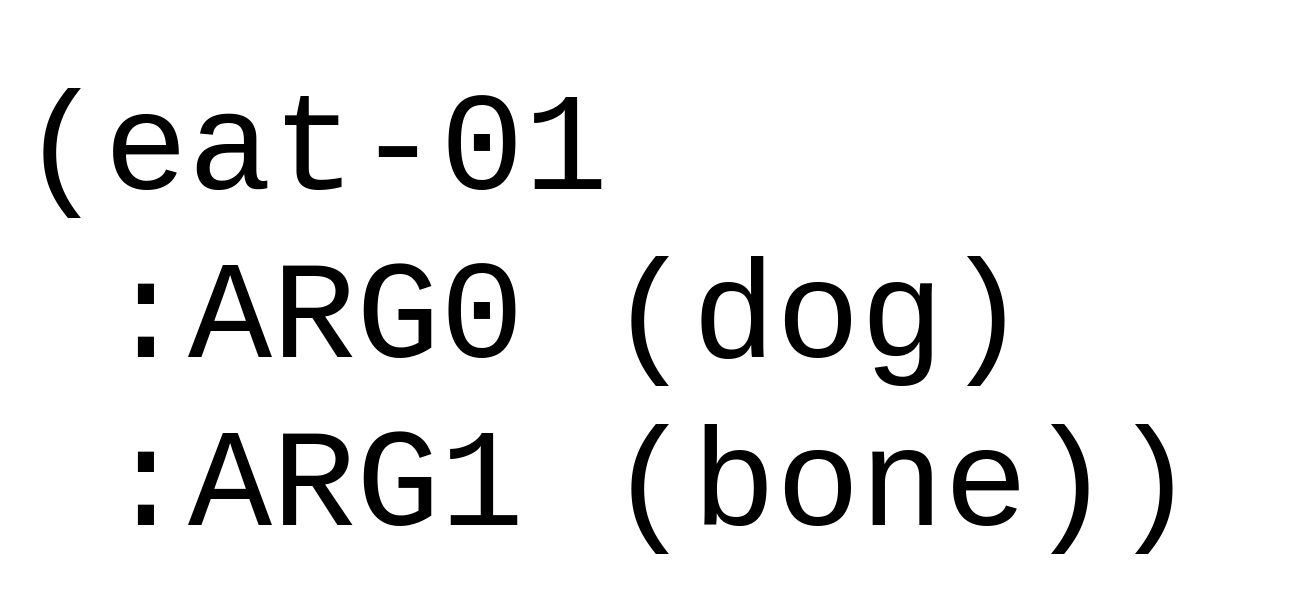} }}%
    \caption{``The dog eats a bone.''}%
    \label{fig:amr_graphs}%
\end{figure}

AMR was originally designed for English texts only. However, \citet{damonte-cohen-2018-cross} demonstrated that AMR could be used for other languages such as Spanish, Italian, Chinese, and German. Since then, many approaches have adopted AMR parsing for multilingual AMR parsing~\cite{procopio-etal-2021-sgl,blloshmi-etal-2020-xl,xu-etal-2021-xlpt,cai-etal-2021-multilingual-amr,sheth-etal-2021-bootstrapping}. However, one of the main challenges for this task is the lack of data. Currently, training data are only available in English \cite{Knight2017,Knight2020} and evaluation data in 6 languages: English, German, Spanish, Italian, Chinese \cite{damonte-cohen-2018-cross,li-et-al2021-amr},\footnote{In Chinese AMR 2.0 \cite{li-et-al2021-amr}, AMR concepts are annotated in Chinese.} and French \cite{kang-etal-2023-analyse}.
To overcome the lack of training data in target languages, previous approaches create silver training data in the target languages. This is done through machine translation \cite{damonte-cohen-2018-cross,blloshmi-etal-2020-xl} under the assumption that a text conveying the same meaning should have a shared AMR graph across languages. Similarly, parallel corpora with English AMR parsers are also employed to create silver data \cite{xu-etal-2021-xlpt,blloshmi-etal-2020-xl}. Another approach uses English data for training and then evaluates the model in the target language in a zero-shot manner \cite{procopio-etal-2021-sgl}. Since evaluation data is available in five languages, most of these proposals focus on this small set of languages. 

In this study, our goal is to apply AMR parsing for more diverse languages that have been less explored in previous work and tackle the lack of training data with $k$-shot learning. Taking inspiration from \citet{langedijk-etal-2022-meta}, who applied meta-learning for $k$-shot cross-lingual syntactic parsing, we apply meta-learning for cross-lingual AMR parsing. To examine the efficiency of the method, we compare the meta-learning approach to a classical joint learning method.

Our contributions to cross-lingual AMR parsing are as follows: 
\begin{itemize}[noitemsep]
  \item This work presents the \textbf{first empirical study on meta-learning applications on cross-lingual AMR parsing}.
  \item We train and evaluate our model in languages less explored for AMR parsing: Korean, Croatian, French, and Farsi.
  \item We \textbf{publish new evaluation data in Korean and Croatian}, based on \textit{The Little Prince}. 
  \item We release a multilingual AMR parser that can be evaluated in many languages in $k$-shot. We also release the code to train and evaluate the model.\footnote{The datasets and codes are both available at \url{https://github.com/Emvista/Meta-XAMR-2024.git}}  
\end{itemize}
 
\section{Meta Crosslingual AMR}

\paragraph{Seq2seq AMR Parsing}

In sequence-to-sequence AMR parsing \cite{Bevilacqua_Blloshmi_Navigli_2021}, AMR parsing is viewed as generating a sequence of tokens representing AMR nodes and edges. AMR graphs should be first linearized in a single-line format (see Figure \ref{fig:amr_graphs}) to feed it to a sequence-to-sequence model. We linearize AMR graphs following  \citet{DBLP:journals/corr/NoordB17a}, which includes light pre-processing such as removing variables and wiki links.\footnote{We employ the implementation code available at \url{https://github.com/RikVN/AMR} for graph preprocessing and postprocessing.} We refer the readers to \citet{DBLP:journals/corr/NoordB17a} for a comprehensive understanding of the linearization process. To generate AMR graphs from multi-lingual inputs, we employ the mBart \cite{tang2020multilingual} model, a pre-trained multilingual sequence-to-sequence model, as done by \citet{procopio-etal-2021-sgl}. 

\begin{figure*}
  \includegraphics[width=\textwidth]{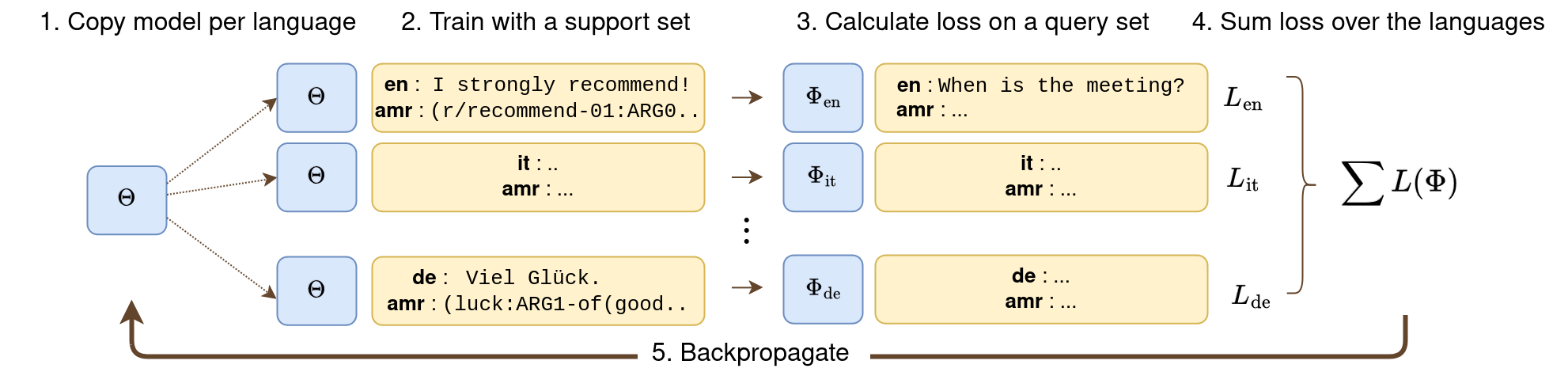}
  \caption{One training step for \textsc{maml} cross-lingual AMR parsing.}
  \label{fig:meta_learning_diagram}
\end{figure*}

\paragraph{MAML for Cross-lingual AMR Parsing}
We use \textsc{maml} \cite{finn2017modelagnostic} for cross-lingual AMR parsing. \textsc{maml} learns good initial parameters $\theta$ that can be tuned to unseen tasks with only a few optimization steps and a few training data examples. \textsc{maml} trains a model to be good at adapting to new tasks only with a few examples by \textit{simulating the $k$-shot training and evaluation} during the training. We apply \textsc{maml}  to train our multilingual AMR parser so that it adapts quickly to new tasks, which are in our case, new languages. The training procedure is described below. 

\noindent \textbf{Step 1}: At each iteration step, the initial model ($\Theta$) is copied once per language $i$. For each $i$, $2 \times K$ examples are randomly sampled from $D_{i}^{\text{train}}$ and divided into the support and the query set ($K$ each). Using the support set, the model is temporarily updated with stochastic gradient descent with learning rate $\alpha$ (Eq.\ \ref{innerloop}). Iterate through the support set for P adaptation steps to obtain $\Phi_{i}$:

\begin{equation}\label{innerloop}
\Phi_{i} \leftarrow \Theta -\alpha \bigtriangledown_{\Theta}\mathcal{L}(\Theta_{i}).
\end{equation}

Next, the loss is computed to evaluate the temporary model $\Phi_{i}$ on the query set. The loss $\mathcal{L}_{i}(\Phi_{i})$ is saved for the next step. The entire step is called an `inner loop' and the inner loop is repeated over the entire task batch, that is, for the number of all training languages $I$.\\

\noindent \textbf{Step 2}: $\mathcal{L}_{i}(\Phi_{i})$ is summed up over training languages to update the initial model $\Theta$ by stochastic gradient descent with a learning rate $\beta$. This entire step is called an `outer loop':\footnote{We apply First-Order \textsc{maml} to avoid computation overhead (second-order derivative requires heavy computation)}

\begin{equation}\label{outerloop}
\Theta \leftarrow \Theta - \beta\sum_{i}\bigtriangledown_{\Phi_{i}} \mathcal{L}_{i}(\Phi_{i}).
\end{equation}

\noindent \textbf{Step 3}: Repeat Step 1 and Step 2 until the total number of training steps.

\section{Experimental Setup}
\label{sec:exp_setup}
\paragraph{Silver Training/Validation Data}

We aim to train a multilingual AMR parser that adapts quickly to new languages, specifically French, Chinese, Korean, Farsi, and Croatian, with $k$ examples. Our method is similar to that of \citet{langedijk-etal-2022-meta} in applying meta-learning for a $k$-shot cross-lingual parsing task, but our training data is only available in English, whereas they have multilingual training data. To create multilingual training data, we apply machine translation as in previous approaches \cite{damonte-cohen-2018-cross,xu-etal-2021-xlpt,blloshmi-etal-2020-xl}. We adopt DeepL\footnote{\url{https://www.deepl.com}} and translate English AMR training data \cite[LDC2020T02]{Knight2020} into 13 languages: German, Italian, Romanian, Finnish, Russian, Turkish, Japanese, Czech, Dutch, Polish, Swedish, Estonian, and Indonesian. The 13 languages were chosen for compatibility with our training model, mBart \cite{tang2020multilingual}, and for language diversity. They cover 5 language families: Indo-European (Germanic, Romance, Slavic), Uralic, Turkic, Japonic, and Austronesian. For each training language, there are 55,635 pairs of sentences and their corresponding AMR graph. To assess the translation quality, we evaluated the training data with the reference-free evaluation metric \textsc{comet} \citep{rei-etal-2020-comet}. The \textsc{comet} score of 13 languages is 83.8$\pm$0.8. We use a total of 14 languages including English for our training data. We use Spanish as the validation language and use the Spanish evaluation set from AMR 2.0~\cite{Damonte2020}. For $k$-shot evaluation during the validation and test step, $k$ random examples from the English dev set are translated to each evaluation language. 

\paragraph{Gold Test Data}
We evaluate our model in French, Chinese, Korean, Farsi, and Croatian. For French, Chinese, and Farsi, we employ \textit{The Little Prince} AMR corpus annotated in each language, respectively from \citet{kang-etal-2023-analyse}, \url{https://amr.isi.edu/} and \citet{takhshid2022persian}.\footnote{The original Farsi dataset consists of AMR concepts in Farsi. Since we employ AMR graphs with English concepts, we use only the input texts of the corpus and graphs from the English AMR corpus.} For Croatian and Korean, we create our test sets by manually aligning \textit{The Little Prince} corpus in each language to corresponding English AMR graphs. After manual alignment, we excluded pairs exhibiting semantic discrepancies between the aligned sentence and its English counterpart, such as pairs where additional or omitted information was observed in the aligned sentences.\footnote{The first author of this article, a native Korean speaker, manually aligned and filtered the data. For Croatian, we automatically translated Croatian text into English with Google Translate (\url{https://translate.google.com/}) and checked the semantic discrepancy with its English counterpart.} This leaves us with, respectively, 1,527 and 1,543 pairs for Korean and Croatian. A few examples of the final dataset are given in Appendix~\ref{appendix:data_samples}. We make the test set publicly available.

\paragraph{Meta-Training and Evaluation} 
We adopt \texttt{mBart-large-50} model \cite{tang2020multilingual} from the transformers library \cite{wolf2020huggingfaces} to train our multilingual AMR parser. To implement model-agnostic meta-learning, we employ the \texttt{learn2learn} library \cite{Arnold2020-ss}. Parameters used for the training are provided in Appendix \ref{appendix:training_parameters}. Our goal is to evaluate the model's performance in new languages that were not seen during the training, specifically, French, Chinese, Korean, Farsi, and Croatian. To this end, for both validation and testing, we employ $k$-shot learning, where the model is fine-tuned with $k$ examples for the test language before evaluation. We report evaluation scores with varying $k$ size using \textsc{smatch}~\cite{cai-knight-2013-smatch}, an evaluation metric for AMR graphs.

\paragraph{Baseline with Joint Learning} 
We train a baseline model with a joint learning method for comparison with our approach. We use the same mBart model and the training data as described above. To assess the effectiveness of our method compared to joint learning, we carry out the two experiments in settings as similar as possible (e.g. training data, hyper-parameters, learning scheduler, $k$-shot evaluation). Hyperparameter details are given in Appendix~\ref{appendix:training_parameters}.

\section{Results and Discussion}

We assessed our model across five languages in $k$-shot learning. Table \ref{tab:smatch_best_model} displays the evaluation results for different shot settings ($k$) where $k={0, 32, 128}$. In the 0-shot evaluation, \textsc{maml} demonstrates higher performance for most evaluation languages, except for Croatian. Nevertheless, the performance gap is minimal, making it difficult to draw firm conclusions regarding the method's advantage. In the $k$-shot evaluation, the performance gap between the two models diminishes, with either the average score showing no significant difference (128-shot) or the baseline model outperforming the \textsc{maml} model (32-shot). These observations suggest that while \textsc{maml} may offer benefits in 0-shot evaluation for certain languages, its advantage is not consistent across all languages. In $k$-shot learning scenarios, the benefit is minimal or null. On the other hand, the joint-learning method shows competitive results regardless of its methodological simplicity. We hypothesize that substantial overlap between inputs and outputs in the training data across languages has contributed to these results. Our training data comprises translations of AMR~3.0 into multiple languages, resulting in overlapped AMR graphs and shared patterns in input texts. In this context, the joint-learning model may learn the similarities between training data directly, allowing the model to learn the task more efficiently. 

Surprisingly, both \textsc{maml} and baseline models exhibit a performance decrease when fine-tuned in 32-shot, compared to not being fine-tuned at all. We hypothesize that the mBart pre-trained model has already enough knowledge of our target languages and fine-tuning the model with only a few examples in each language may impair the model's capacity. This could also be attributed to the domain difference between the fine-tuning dataset and the test dataset. The fine-tuning dataset includes content from general fields such as online forums, journals, and web blogs, whereas the test dataset consists of \textit{The Little Prince}, a novel written in the 1940s. Consequently, the domain shift between the two datasets may have contributed to the model's inability to generalize effectively to the test domain. 

We provide additional analysis of our models in Appendix~\ref{appendix:additional_analysis} (effect of the number of considered languages and of the translation quality).

\section{Related Work}

Meta-learning, also known as \textit{learning to learn}, is a learning paradigm that allows a model to quickly learn a new task with only a few examples. This is made possible by the prior knowledge that the model has acquired through a series of different tasks. In cross-lingual applications, each task corresponds to a different language. The closest approach to ours is \citet{langedijk-etal-2022-meta}, who adopt \textsc{maml} for cross-lingual dependency parsing. They train a dependency parser on a set of languages using \textsc{maml} and then evaluate the model on unseen languages to investigate the model's ability to adapt quickly. In contrast, we focus on a \textit{semantic} parsing task with an unanchored formalism. In addition, they have multilingual training data at hand, whereas we generate our silver multilingual data by machine translation from English data. Another difference is that they use a graph-based biaffine model for parsing, whereas we use a seq2seq model with a linearized graph. \citet{sherborne-lapata-2023-meta} applied meta-learning to cross-lingual SQL parsing. While useful at representing (and executing) database queries expressed in natural language, SQL is not a general-purpose semantic formalism like AMR. To the best of our knowledge, our work is the first to apply \textsc{maml} for cross-lingual AMR parsing.

\begin{table}[]
\center
\resizebox{\columnwidth}{!}{
\begin{tabular}{@{}lllllll@{}}
\toprule
 & fr & zh & ko & fa & hr & avg \\ \midrule
base\_0-shot & 56.4 & 45.6 & 42.1 & 46.3 & \textbf{51.4} & 48.4 \\
\textsc{maml}\_0-shot & \textbf{56.5} & \textbf{46.1} & \textbf{42.2} & \textbf{46.7} & 50.8 & \textbf{48.5} \\ \midrule
base\_32-shot & \textbf{56.3} & \textbf{45.4} & \textbf{42.0} & \textbf{46.1} & \textbf{51.3} & \textbf{48.3} \\
\textsc{maml}\_32-shot & 55.5 & 45.1 & 41.1 & 45.9 & 48.9 & 47.3 \\ \midrule
base\_128-shot & \textbf{56.5} & 45.9 & 42.0 & 46.6 & \textbf{51.5} & 48.5 \\
\textsc{maml}\_128-shot & 56.0 & \textbf{46.2} & \textbf{42.2} & \textbf{46.8} & 51.3 & 48.5 \\ \bottomrule
\end{tabular}
}
\caption{\textsc{smatch} scores of the baseline and the \textsc{maml} model ($k$-shot evaluation).}
\label{tab:smatch_best_model}
\end{table}

\section{Conclusion}
This study investigates the effectiveness of meta-learning compared to joint learning in cross-lingual AMR parsing. We assess our models across less-explored languages for AMR parsing, including French, Chinese, Korean, Farsi, and Croatian. To facilitate evaluation, we develop new test sets for Korean and Croatian and release the data to promote AMR parsing in diverse languages. Our findings reveal that meta-learning exhibits minor performance gain compared to joint learning in 0-shot evaluation. The small gain diminishes for $k$-shot learning (when $k>0$). Consequently, our results suggest that the joint learning method serves as a robust baseline, while meta-learning appears to be a sub-optimal approach for cross-lingual AMR parsing. We believe that this research provides valuable insights into the comparative efficacy of meta-learning and joint learning in cross-lingual AMR parsing, offering important guidance for future developments in cross-lingual AMR parsers. 

\section*{Limitations}

Our model does not outperform a simple monolingual model which is trained with AMR data in the target language translated by a MT system. However, our approach can be explored for low-resource languages for which machine translation is not available. In addition, we did not apply grid search to find the best learning rates for the baseline models and used the same learning rate as done by \citet{procopio-etal-2021-sgl}, who also employed mBart for sequence-to-sequence cross-lingual AMR parsing. This could have affected the results in favor of meta-learning. Nonetheless, this does not affect our conclusion of the empirical study to reveal the weakness of the meta-learning approach for cross-lingual AMR parsing. This study does not include evaluation scores on the AMR 2.0 multilingual test set, which could help position our models relative to the state-of-the-art models. There are two motivations for the omission. Firstly, the Spanish test set in AMR 2.0 is already used as our validation set. Therefore, the AMR graphs (they are shared across the 4 languages) are already exposed during the validation step. Secondly, German and Italian, evaluation languages in AMR 2.0, are already included in our training data. Since our goal is to evaluate our model for unseen target tasks, evaluating our model on these languages is not coherent with the objective. Despite the limitations, we believe that our study empirically shows the constraints of meta-learning for cross-lingual AMR parsing and provides valuable insights into the meta-learning application in the task. 

\section*{Acknowledgement}
We gratefully acknowledge the insightful comments and thoughtful suggestions provided by the anonymous reviewers. This work was granted access to the HPC resources of IDRIS under the allocation 2024-AD011012853R2 made by GENCI.

\bibliography{custom}
\bibliographystyle{acl_natbib}
\label{sec:bibtex}

\newpage

\appendix

\section{Aligned Data Samples}
\label{appendix:data_samples}
\noindent\fbox{%
    \parbox{\columnwidth}{%
        \textbf{en} In the book it said : " Boa constrictors swallow their prey whole , without chewing it .\\
\textbf{ko} 그 책에는 이렇게 씌어 있었다. "보아 구렁이는 먹이를 씹지도 않고 통째로 집어삼킨다\\
\textbf{hr} U knjizi je pisalo: »Udavi gutaju svoj plijen cijel cjelcat, bez žvakanja. \\

\textbf{en} I pondered deeply , then , over the adventures of the jungle .\\
\textbf{ko} 나는 그래서 밀림 속에서의 모험에 대해 한참 생각해 봤다. \\
\textbf{hr} Zatim sam mnogo razmišljao o prašumskim pustolovinama,\\

\textbf{en} The little prince , who asked me so many questions , never seemed to hear the ones I asked him . \\
\textbf{ko} 어린 왕자는 내게 많은 것을 물어보면서도 내 질문에는 귀를 기울이는 것 같지 않았다.\\
\textbf{hr} Činilo se da mali princ, koji mi je postavljao brojna pitanja, nikada ne čuje moja. \\

\textbf{en} I was more isolated than a shipwrecked sailor on a raft in the middle of the ocean . \\
\textbf{ko} 대양 한가운데에 떠 있는 뗏목 위의 표류자보다 나는 더 고립되어 있었다. \\
\textbf{hr} Bio sam usamljeniji od brodolomca na splavi usred oceana.
    }%
}

\section{Training Hyperparameters}
\label{appendix:training_parameters}

\paragraph{Meta Crosslingual AMR} We train our model for 30,000 steps and evaluate the model every 500 steps with the Spanish validation set. Early stopping is applied, terminating training if the dev \textsc{smatch} score fails to improve for more than 7,500 steps. The number of fine-tuning cycles, called an adaptation step, is denoted as $P$. Unless specified otherwise, we set $P=0$ and $k=0$ (0-shot learning). \textsc{maml} requires two learning rates, one for the inner loop ($\alpha$) and one for the outer loop ($\beta$). We conducted a grid search to identify an optimal learning rate set and used $\alpha = \num{1e-5}$, $\beta = \num{3e-5}$ throughout the experiments. For $\beta$, we use a linear learning rate scheduler with 1,500 warm-up steps. Unless specified otherwise, we apply \num{1e-5} to fine-tune a model before validation/testing. At each iteration step during the training, $2 \times K$ are sampled to form a query and a support set for each training language. As a result, the batch size $N$ equals $2 \times K \times I$, where $I$ denotes the number of training languages. By default, we assign $K=8$ and $I=14$, unless stated otherwise.

\paragraph{Baseline Model} 
For the training set, we use a concatenation of the multilingual AMR training sets described in Section~\ref{sec:exp_setup}. At each iteration step, we randomly select $N$ training examples from the concatenated training sets to calculate the loss and optimize the model accordingly. For the rest of the hyperparameters and test/evaluation method, we apply the same settings as described as above (e.g. learning rate scheduler, $k$-shot size) except for the learning rate since maml requires two learning rates $\alpha$ and $\beta$ whereas joint-learning requires only one. We use a uniform learning rate for training \num{3e-5} with a linear scheduler with 1500 warm-up steps.

\section{Additional Analysis}
\label{appendix:additional_analysis}
We provide additional analysis of our approach focusing on how the training is affected by the number of training languages and translation sources. The results include 0-shot evaluation for both meta-learning and joint learning. 

\subsubsection*{Q1: How does the number of languages affect the performance of the models? }

To examine how the number of training languages impacts the model performance, we incrementally add more languages to the training data and we train three models respectively with 8, 12, and 14 languages. The first model is trained in German, English, Italian, Romanian, Russian, Turkish, Finnish, and Japanese. Then we add Czech, Dutch, Polish, and Swedish, and then finally we add Estonian and Indonesian. Note that for meta-learning, the batch size depends on the number of training tasks since we randomly sample $K$ examples per language ($\text{batch size} = 2 \times K \times I $ where $I$ denotes the number of training languages). To keep the batch size consistent across experiments while altering only the number of languages, when more than 8 languages are used for training, we randomly sample 8 languages per iteration step and select $K$ training examples per language. Unless specified otherwise, each model is evaluated in a zero-shot manner for five languages: French, Chinese, Korean, Farsi, and Croatian. 

\paragraph{Results} Table~\ref{tab:smatch_lang} shows that both the \textsc{maml} and baseline models have a positive correlation with the number of training languages.  
The baseline model has the largest gain when increasing the number of languages from 8 to 12 language by 15.7\%. \textsc{Maml} models, on the other hand, have the biggest gain when increasing the number of languages from 12 to 14 languages by 14.2\%. Looking in detail per target language, however, in the \textsc{maml} model, not all target languages benefit from adding more training languages. Comparing the two \textsc{maml} models, trained respectively with 8 languages and 12 languages, the \textsc{smatch} score drops in Chinese and Farsi when adding four languages to the training data, whereas the baseline model shows a steady increase across target languages when adding more languages. In other words, the baseline model benefits uniformly from the inclusion of more training languages across all target languages, while the performance of the \textsc{maml} model varies depending on the specific target language. In the \textsc{maml} models, certain languages experience a decrease in performance despite the addition of more training languages. 
A caveat of this experiment is that the results may depend on the order in which the languages are added and their typological relationship to evaluation languages (we leave this investigation to future work).

\begin{table}[]
\resizebox{\columnwidth}{!}{
\begin{tabular}{@{}lllllll@{}}
\toprule
 & fr & zh & ko & fa & hr & avg \\ \midrule
base\_14langs & 56.3 & 45.6 & 42.1 & 46.3 & 51.4 & 48.4 \\
base\_12langs & 53.6 & 41.6 & 40.1 & 43.4 & 45.9 & 44.9 \\
base\_8langs & 47.5 & 39.8 & 39.1 & 40.5 & 22.4 & 37.8 \\
\midrule
\textsc{maml}\_14langs & 56.5 & 46.1 & 42.2 & 46.7 & 50.8 & 48.5 \\
\textsc{maml}\_12langs & 48.5 & 39.4 & 35.1 & 39.7 & 45.0 & 41.5 \\
\textsc{maml}\_8langs & 47.7 & 39.6 & 34.3 & 40.1 & 42.4 & 40.8 \\ \bottomrule
\end{tabular}
}
\caption{\textsc{smatch} scores according to the number of training languages.}
\label{tab:smatch_lang}
\end{table}

\subsubsection*{Q2: How robust is the model with respect to translation quality?}
To assess the impact of the translation source on our method, we employ an alternative translation model to translate our training data. Specifically, we use the mBart translation models, sourced from the Huggingface hub\footnote{\url{https://huggingface.co/facebook/mbart-large-50-many-to-many-mmt}}, to translate our training data into 13 languages. \textsc{comet} score of the 13 translated texts is 80.7$\pm$1.4. Subsequently, we use this translation to train both the \textsc{maml} and baseline models. Following this, we contrast the evaluation outcomes of these models with those trained using the DeepL translation.

\paragraph{Results} For both the \textsc{maml} and the baseline models, when using an open-source translation model mBart, the performance drops (see Table~\ref{tab:smatch_translation}). In both cases, the Korean \textsc{smatch} score drops the most when using the mBart translation model. \textsc{maml} model is more affected by this change. On the average score, the baseline model drops by 0.9\%, whereas the \textsc{maml}-model drops by 2.3\%. This result shows that the meta-learning model is more sensitive to the input texts than the baseline model.

\begin{table}[]
\resizebox{\columnwidth}{!}{
\begin{tabular}{@{}lllllll@{}}
\toprule
 & fr & zh & ko & fa & hr & avg \\ \midrule
base\_DeepL & 56.3 & 45.6 & 42.1 & 46.3 & 51.4 & 48.4 \\
base\_mBart & 56.2 & 44.5 & 41.2 & 46.1 & 51.3 & 47.8 \\
\midrule
\textsc{maml}\_DeepL & 56.5 & 46.1 & 42.2 & 46.7 & 50.8 & 48.5 \\
\textsc{maml}\_mBart & 55.6 & 45.1 & 40.8 & 46.1 & 48.9 & 47.3 \\ \bottomrule
\end{tabular}
}
\caption{\textsc{smatch} scores according to the translation source.}
\label{tab:smatch_translation}
\end{table}

\end{document}